\documentclass[letterpaper, 10 pt, conference]{ieeeconf}  

\IEEEoverridecommandlockouts                              

\usepackage{multirow}
\usepackage{bbding}
\usepackage{pifont}
\usepackage[nointegrals]{wasysym}
\usepackage{caption}
\usepackage{subcaption}
\usepackage{amsmath}
\usepackage{array}
\usepackage{color}
\usepackage[normalem]{ulem}
\usepackage{hyperref}
\usepackage{nomencl}
\makenomenclature
\usepackage[ruled,vlined]{algorithm2e}
\usepackage{amsmath,amssymb,amsfonts}
\usepackage{algorithmic}
\usepackage{graphicx}
\usepackage{textcomp}
\usepackage{wrapfig}
\usepackage{mdframed,lipsum}
\usepackage[T1]{fontenc} 
\usepackage{booktabs}
\usepackage{mwe}
\usepackage[table, dvipsnames]{xcolor}
\usepackage{array,booktabs}
\newcommand{\STAB}[1]{\begin{tabular}{@{}c@{}}#1\end{tabular}}
\usepackage{cite}

\usepackage{caption}
\SetKwInput{KwInput}{Input}
\SetKwInput{KwOutput}{Output}

\title{\LARGE \bf
Enhancing State Estimation in Robots: A Data-Driven Approach with Differentiable Ensemble Kalman Filters
}

\author{Xiao Liu$^{1}$, Geoffrey Clark$^{1}$, Joseph Campbell$^{2}$, Yifan Zhou$^{1}$, and Heni Ben Amor$^{1}$%
\thanks{$^{1}$X.~Liu, G.~Clark, Y.~Zhou, and H.~Ben~Amor are with SCAI at Arizona State University, USA {\tt\small \{xliu330, gmclark1, yzhou298, hbenamor\}@asu.edu}}%
\thanks{$^{2}$J.~Campbell is with the RI at Carnegie Mellon University, USA {\tt\small jcampbell@cmu.edu}}%
}

\begin{document}

\maketitle
\thispagestyle{empty}
\pagestyle{empty}

\begin{abstract}

This paper introduces a novel state estimation framework for robots using differentiable ensemble Kalman filters (DEnKF). DEnKF is a reformulation of the traditional ensemble Kalman filter that employs stochastic neural networks to model the process noise implicitly. Our work is an extension of previous research on differentiable filters, which has provided a strong foundation for our modular and end-to-end differentiable framework. This framework enables each component of the system to function independently, leading to improved flexibility and versatility in implementation. Through a series of experiments, we demonstrate the flexibility of this model across a diverse set of real-world tracking tasks, including visual odometry and robot manipulation. Moreover, we show that our model effectively handles noisy observations, is robust in the absence of observations, and outperforms state-of-the-art differentiable filters in terms of error metrics. Specifically, we observe a significant improvement of at least 59\% in translational error when using DEnKF with noisy observations. Our results underscore the potential of DEnKF in advancing state estimation for robotics. Code for DEnKF is available at \url{https://github.com/ir-lab/DEnKF}
\end{abstract}

\section{INTRODUCTION}
In robotics, Recursive Bayesian filters, especially Kalman filters, play a crucial role in accurately localizing robots in their surroundings~\cite{thrun2005probabilistic}, predicting the future movements of human interaction partners~\cite{WANG2022102310}, tracking objects over time~\cite{chen2011kalman}, and ensuring stability during robot locomotion~\cite{reher2019dynamic}. Typically, the success of these filters depends on having an accurate model of the dynamics of the system being observed, as well as a model of the observation process itself. However, modeling complex systems and their noise profiles can be a challenging task, often requiring additional steps like statistical modeling or system identification. Despite advancements in this field, such as particle filters~\cite{thrun2002probabilistic}, scalability remains an issue when working with high-dimensional systems.

The limitations of traditional Bayesian filters have inspired the development of Deep State-Space Models (DSSM)~\cite{NEURIPS2018_5cf68969, klushyn2021latent, kloss2021train}. DSSM leverages deep learning techniques to learn approximate, nonlinear models of the underlying states and measurements from recorded data. This approach aims to overcome the need for explicit modeling of the processes, which can be challenging for complex dynamical systems. However, incorporating the nonlinear capabilities of neural networks into recursive filtering may come with additional linearization steps or limitations~\cite{thrun2005probabilistic}, which can impact the quality of the inference.
\begin{figure}[t!]
\centering
\includegraphics[width=\linewidth]{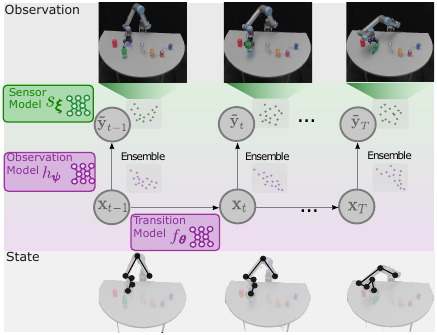}
\caption{Differentiable Ensemble Kalman Filter (DEnKF), employs an ensemble state to represent the probability density and integrates a stochastic neural network to generate state transitions. In order to project high-dimensional visual inputs to observation space, a sensor model is utilized. By combining an ensemble Kalman Filter with a learned observation model, the DEnKF is able to achieve precise posterior state estimations.}
\label{fig:overview}
\vspace{-0.2in}
\end{figure} 
In this paper, we introduce a new approach to robot state estimation by extending prior research on differentiable recursive filters. Specifically, we propose the Differentiable Ensemble Kalman Filter (DEnKF), which employs high-dimensional camera images to estimate and correct the state of a robot arm, as demonstrated in Fig.~\ref{fig:overview}. Our method builds upon the solid foundation established by prior works on differentiable filters, including the contributions made in~\cite{klushyn2021latent, kloss2021train, jonschkowski2018differentiable}. Our approach addresses the challenges encountered by differentiable filtering through both theoretical and practical innovations. One such innovation involves the sampling of states from the posterior distribution of a neural network, which eliminates the need to estimate noise parameters for the recursive filter. Another innovation involves the ensemble formulation of the filtering process, which eliminates the need for linearization. Notably, unlike many other Deep State-Space Models (DSSM) in the literature, our approach avoids the use of Recurrent Neural Networks (RNNs), which have been shown to limit the accuracy of learned models and may lead to non-Markovian state-spaces~\cite{klushyn2021latent}.

In this paper, we present an end-to-end learning approach for recursive filtering that simultaneously learns the observation, dynamics, and noise characteristics of a robotic system. The key contributions of our work can be summarized as follows:
\begin{itemize}
    \item A stochastic state transition model that uses samples from the posterior of a neural network to implicitly model the process noise, avoiding the need for a parametric representation of the posterior.
        
    \item An ensemble formulation that allows for the efficient inference of both linear and nonlinear systems, without the need for an explicit covariance matrix, making it suitable for high-dimensional inputs and noisy observations.
    
    \item Empirical evaluations for the autonomous driving task show DEnKF effectively reduce the translational and rotational errors compared to state-of-the-art methods, reducing errors by up to 59\% and 36\% when dealing with noisy observations, and handling missing observation scenarios with improved error reductions by 2-fold and 3-fold.

    

\end{itemize}

\section{Related work}
Kalman filters (KFs) are well-studied and widely-used state-space models with many applications in robotics~\cite{thrun2005probabilistic}. KFs are designed for systems with linear process and observation models and assume normally distributed noise. To overcome some of these limitations and extend the inference capabilities to nonlinear systems, several variants have been proposed, e.g., the Extended Kalman Filter (EKF)~\cite{Sorenson} and the Unscented Kalman Filter (UKF)~\cite{van2004sigma}. 
Still, even the EKF and UKF face theoretical and computational challenges when dealing with high-dimensional observations. Among the many reasons for this limitation is the need for explicitly calculating an error covariance during the filtering process.

\textbf{Differentiable filters}: Differentiable filters (DFs) aim to adapt the recursive filtering techniques to handle high-dimensional inputs. For instance, BackpropKF~\cite{haarnoja2016backprop} proposed a way to train Kalman Filters as recurrent neural networks using backpropagation with the integration of feed-forward networks and convolutional neural networks. Similarly, Differentiable algorithm networks~\cite{karkus2019differentiable} introduced neural network components that encode differentiable robotic algorithms. This methodology is similar to that of Differentiable Particle Filters (DPFs)\cite{jonschkowski2018differentiable, corenflos2021differentiable,chen2021differentiable}, which employ algorithmic priors to increase learning efficiency. Variations of DPFs were explored in~\cite{wang2019dualsmc} and~\cite{wen2021end} using adversarial methods for posterior estimation or partial ground truth particles for semi-supervised learning. The training of DFs and modeling of uncertainty along with noise profiles were analyzed in~\cite{kloss2021train}. The authors implemented the components of the DFs as multi-layer perceptrons and enveloped the sub-modules in an RNN layer. 
The DFs in~\cite{kloss2021train,lee2020multimodal} were tested on real-world tasks, indicating that end-to-end learning is crucial for learning accurate noise models. Similarly,~\cite{wagstaff2022self} developed a self-supervised visual-inertial odometry model using the differentiable Extended Kalman Filter (dEKF) based on the work in~\cite{kloss2021train}. 
 However, as noted in~\cite{klushyn2021latent}, RNNs can often be a limiting factor in learning accurate models of the system dynamics. The current lack of DFs that handle missing observations is attributed to the use of RNN frameworks to model system dynamics.

\textbf{Ensemble Kalman Filters}: A modern variant of KFs that is particularly successful on high-dimensional, nonlinear tasks is Ensemble Kalman Filters (EnKFs)~\cite{evensen2003ensemble}. EnKFs have been shown to enable accurate estimation of state-space dynamics in data assimilation tasks~\cite{roth2017ensemble} without linearity assumptions. They have found popularity in modeling and forecasting complex weather phenomena that may include millions of state dimensions~\cite{houtekamer2016review}. Rather than assuming a certain parametric form of the underlying distribution, EnKFs approximate the posterior distribution through an ensemble (or collection) of state vectors. They are computationally efficient since they do not require the explicit calculation (and inversion) of error covariance matrices. In addition, EnKFs do not require explicit parametric characterizations of the process and observation noise. Instead, it only requires the ability to generate samples from the underlying distributions. 

In this paper, we leverage this \textbf{key insight} in order to create a theoretical and practical connection between EnKFs and stochastic Bayesian neural networks (SNNs)~\cite{jospin2022, lakshminarayanan2017simple}. 
As stated in~\cite{chua2018deep}, SNNs can model two types of uncertainty: 1) aleatoric uncertainty, which arises from inherent stochasticities of a system, i.e., process noise and observation noise; 2) epistemic uncertainty, which is caused by a lack of sufficient data to determine the underlying system uniquely. The integration between EnKFs and SNNs results in a probabilistic filtering process that leverages the advantages of modern neural network techniques. 
First attempts at differentiable EnKFs were proposed in~\cite{chen2021auto} which searches for optimal parameters utilizing gradient information from EnKFs. 
This differs substantially from our approach, which is a fully differentiable end-to-end framework. Note that EnKF is theoretically related to Particle Filters (PFs) -- both are Monte-Carlo filtering techniques based on similar principles. However, in contrast to PFs, EnKFs provide equal weight to each ensemble member thereby eschewing the well-known sample degeneracy problem. In addition, EnKFs have also been shown to efficiently model complex phenomena using relatively small ensemble sizes~\cite{zhang2007system}. A study of the approximation error of these filters~\cite{quang2016high} also indicates that with increasing size of state dimensions, EnKFs show a much slower rate of degradation than the PFs. 

\section{Differentiable EnKF}
Recursive Bayesian filtering addresses the general challenge of estimating the state ${\bf x}_t$ of a discrete-time dynamical system given a sequence of noisy observations ${\bf y}_{1:t}$. The posterior distribution of the state can be represented as:
\begin{equation}
\begin{aligned}
\label{eq:01}
    p({\bf x}_t | {\bf y}_{1:t}, {\bf x}_{1:t-1}) &\propto
    p({\bf y}_{t} | {\bf x}_t)\  p({\bf x}_t | {\bf y}_{1:t-1}, {\bf x}_{1:t-1}).
\end{aligned}
\end{equation}
Let $\text{bel}({\bf x}_t) = p({\bf x}_t | {\bf y}_{1:t}, {\bf x}_{1:t-1})$, applying the Markov property, i.e., the assumption that the next state is dependent only upon the current state, yields:
\begin{equation}
\begin{aligned}
\label{eq:02}
    \text{bel}({\bf x}_t) = \underbrace{p({\bf y}_{t} | {\bf x}_t)}_{\text{observation model}}
    \prod_{t=1}^t \overbrace{ p({\bf x}_t|{\bf x}_{t-1})}^{\text{state transition model}} \text{bel}({\bf x}_{t-1}),
\end{aligned}
\end{equation}
where $p({\bf y}_{t} | {\bf x}_t)$ is the observation model and $p({\bf x}_t|{\bf x}_{t-1}$) is the transition model. The transition model describes the laws that govern the evolution of the system state. By contrast, the observation model identifies the relationship between the hidden, internal state of the system and observed, noisy measurements.
An alternative approach of KFs and its variants is to leverage modern deep learning techniques in order to extract complex, nonlinear transition and observation models. Starting with the state transition model in linear KFs:
\begin{equation}
    \begin{aligned}\label{eq:03}
    {\bf x}_t = {\bf A} {\bf x}_{t-1} + {\bf q}_t \quad
    {\bf q}_t \thicksim \mathcal{N}(0, {\bf Q}_{t}).
    \end{aligned}
\end{equation}
the work in~\cite{kloss2021train} replaces the transition matrix ${\bf A}$ and the process noise ${\bf Q}_{t}$ with trained neural networks $f_{\pmb {\theta}}$ and $q_{\pmb {\psi}}$\ respectively:
\begin{equation}
    \begin{aligned}\label{eq:04}
    {\bf x}_t = f_{\pmb {\theta}}({\bf x}_{t-1}) + \mathcal{N}\left(0, q_{\pmb {\psi}} ({\bf{x}}_{t-1}) \right),
    \end{aligned}
\end{equation}
where ${\pmb {\theta}}$ and ${\pmb {\psi}}$ denote the neural network weights.
Note, that the network $q_{\pmb {\psi}}(\cdot)$ produces the entries of the covariance matrix ${\bf Q}_t$ representing a Gaussian distribution. As shown in Eq.~\ref{eq:04}, the state estimate ${\bf x}_t$ is calculated by generating a sample from a normal distribution with covariance ${\bf Q}_t$ which is then added to the neural network prediction $f_{\pmb {\theta}}({\bf x}_{t-1})$. 
Hence, the process model and the process noise are calculated using two separate neural networks which may not be producing outputs consistent with each other.   


\subsection{Stochastic Neural Models of Dynamics \label{sec:bnn}}
In this paper, we avoid this separation by using recent insights in stochastic neural networks (SNNs)~\cite{jospin2022}. More specifically, the work in~\cite{gal2016dropout} has established a theoretical link between the Dropout training algorithm and Bayesian inference in deep Gaussian processes. Accordingly, after training a neural network with Dropout, it is possible to generate empirical samples from the predictive posterior via \emph{stochastic forward passes}.  
Hence, for the purposes of filtering, we can \textbf{implicitly model the process noise} by sampling state from a neural network trained on the transition dynamics, i.e., ${\bf x}_{t}  \thicksim  f_{\pmb {\theta}} ({\bf x}_{t-1})$. In contrast to previous approaches, the transition network $f_{\pmb {\theta}}(\cdot)$ models the system dynamics, as well as the inherent noise model in a consistent fashion without imposing diagonality.

\subsection{Nonlinear Filtering with Differentiable Ensembles}
Introducing non-linearities through neural network realizations of the transition and observation function invalidates the linearity assumptions that are the backbone of many recursive Bayesian filters. To overcome this challenge, we embed our methodology within an EnKF framework. Throughout the filtering process, each ensemble member is propagated forward in time to yield a new approximate posterior distribution. EnKF does not require an explicit representation of the process and observation noise -- instead we only need to be able to sample from the noise distribution. We formulate DEnKF as an extension of the EnKF while keeping the core algorithmic steps intact. 
In particular, we use an initial ensemble of $E$ members to represent the initial state distribution ${\bf X}_0 = [ {\bf x}^{1}_0, \dots, {\bf x}^{E}_0]$, $E \in \mathbb{Z}^+$.

\textbf{Prediction Step}: We leverage the stochastic forward passes from a trained state transition model to update each ensemble member: 
    \begin{equation}
    \begin{aligned}\label{eq:1}
          {\bf x}^{i}_{t|t-1} & \thicksim  f_{\pmb {\theta}} ({\bf x}^{i}_{t|t-1}|{\bf x}^{i}_{t-1|t-1}),\  \forall i \in E.
    \end{aligned}
   \end{equation}
 Matrix ${\bf X}_{t|t-1} = [{\bf x}^{1}_{t|t-1}, \cdots, {\bf x}^{E}_{t|t-1}]$ holds the updated ensemble members which are propagated one step forward through the state space. Note that sampling from the transition model $f_{\pmb {\theta}}(\cdot)$ (using the SNN methodology described above) implicitly introduces a process noise.

\textbf{Update Step}: Given the updated ensemble members ${\bf X}_{t|t-1}$, a nonlinear observation model $h_{\pmb {\psi}}(\cdot)$ is applied to transform the ensemble members from the state space to observation space. Following our main rationale, the observation model is realized via a neural network with weights $\pmb {\psi}$. Accordingly, the update equations for the EnKF become:
    \begin{align}
    \label{eq:2}
        {\bf H}_t {\bf X}_{t|t-1} &= \left[ h_{\pmb {\psi}}({\bf x}^1_{t|t-1}), \cdots, h_{\pmb {\psi}}({\bf x}^E_{t|t-1}) \right],\\
        \label{eq:3}
        {\bf H}_t {\bf A}_{t} &=  {\bf H}_t {\bf X}_{t|t-1} \\
        &- \left[\frac{1}{E} \sum_{i=1}^E h_{\pmb {\psi}}({\bf x}^i_{t|t-1}),
        \cdots,
        \frac{1}{E} \sum_{i=1}^E h_{\pmb {\psi}}({\bf x}^i_{t|t-1})\right]. \nonumber
    \end{align}
${\bf H}_t {\bf X}_{t|t-1}$ is the predicted observation, and ${\bf H}_t {\bf A}_{t}$ is the sample mean of the predicted observation at $t$. EnKF treats observations as random variables. Hence, the ensemble can incorporate a measurement perturbed by a small stochastic noise thereby accurately reflecting the error covariance of the best state estimate~\cite{evensen2003ensemble}. In our differentiable version of the EnKF, we also incorporate a sensor model which can learn projections between a latent space and higher-dimensional observations spaces, i.e. images. To this end, we leverage the methodology from Sec.~\ref{sec:bnn} to train a stochastic sensor model $s_{\pmb {\xi}}(\cdot)$:
    \begin{equation}
    \begin{aligned}\label{eq:sensor}
          \tilde{{\bf y}}^{i}_t & \thicksim  s_{\pmb {\xi}} (\tilde{{\bf y}}^{i}_t|{\bf y}_{t}),\  \forall i \in E.\\
    \end{aligned}
   \end{equation}
where ${\bf y}_{t}$ represents the noisy observation. Sampling yields observations $\tilde{{\bf Y}}_t = [\tilde{{\bf y}}^{1}_t, \cdots, \tilde{{\bf y}}^{E}_t]$ and sample mean $\tilde{{\bf y}}_t = \frac{1}{E}\sum_{i=1}^i\tilde{{\bf y}}^i_t$.  The innovation covariance ${\bf S}_t$ can then be calculated as:
    \begin{equation}
    \begin{aligned}\label{eq:4}
        {\bf S}_t &= \frac{1}{E-1}  ({\bf H}_t {\bf A}_t)  ({\bf H}_t {\bf A}_t)^T + r_{\pmb {\zeta}}(\tilde{{\bf y}_t}).
    \end{aligned}
    \end{equation}
where $r_{\pmb {\zeta}}(\cdot)$ is the measurement noise model implemented using MLP. We use the same way to model the observation noise as in~\cite{kloss2021train}, $r_{\pmb {\zeta}}(\cdot)$ takes an learned observation $\tilde{{\bf y}_t}$ in time $t$ and provides stochastic noise in the observation space by constructing the diagonal of the noise covariance matrix. The final estimate of the ensemble ${\bf X}_{t|t}$ can be obtained by performing the measurement update step:
    \begin{align}
    \begin{split}\label{eq:5}
        {\bf A}_t &= {\bf X}_{t|t-1} - \frac{1}{E}\sum_{i=1}^E{\bf x}^i_{t|t-1},
    \end{split}\\
    \begin{split}\label{eq:6}
     {\bf K}_t &= \frac{1}{E-1} {\bf A}_t ({\bf H}_t {\bf A}_t)^T {\bf S}_t^{-1},
    \end{split}\\
    \begin{split}\label{eq:7}
    {\bf X}_{t|t} &= {\bf X}_{t|t-1} + {\bf K}_t (\tilde{{\bf Y}}_t - {\bf H}_t {\bf X}_{t|t-1}),
    \end{split}
    \end{align}
where ${\bf K}_t$ is the Kalman gain. In inference, the ensemble mean ${\bf \bar{x}}_{t|t} = \frac{1}{E}\sum_{i=1}^E {\bf x}^i_{t|t}$ is used as the updated state. The neural network structures for all learnable modules are described in Table~\ref{tab:EnKF_module}. Furthermore, we highlight couple of the theoretical properties (in Appendix) of EnKF and its relations to DEnKF. 

\begin{table}[h!]
  \centering
  \caption{Differentiable EnKF learnable sub-modules.}
  \label{tab:EnKF_module}
  \scalebox{0.90}{
  \begin{tabular}{ll}
    \toprule
$f_{\pmb {\theta}}$: & 2$\times$SNN(32, ReLu), 2$\times$SNN(64, ReLu), 1$\times$SNN(S, -)\\
$h_{\pmb {\psi}}$: & 2$\times$fc(32, Relu), 2$\times$fc(64, ReLu), 1$\times$ fc(O, -)\\
$r_{\pmb {\zeta}}$: & 2$\times$fc(16, ReLu), 1$\times$fc(O, -)\\


\multirow{3}{1em}{$s_{\pmb {\xi}}$:} & conv(7$\times$7, 64, stride 2, ReLu), conv(3$\times$3, 32, stride 2, ReLu), \\
 & conv(3$\times$3, 16, stride 2, ReLu), flatten(), 2$\times$SNN(64, ReLu),  \\
 & 2$\times$SNN(32, ReLu), 1$\times$SNN(O, -)\\
    \bottomrule
\multicolumn{2}{l}{fc: fully connected, conv: convolution, S, O: state and observation dimension.} \\
  \end{tabular}}
  \vspace{-0.2in}
\end{table}

\section{Experiments}\label{sec:exp}
We evaluate the DEnKF framework on two common robotics tasks:  a) a visual odometry task for autonomous driving and b) a robot manipulation task in both simulation and real-world. We compare our results to a number of state-of-the-art differential filtering methods~\cite{kloss2021train, jonschkowski2018differentiable, haarnoja2016backprop}.  

\textbf{Training:} DEnKF contains four sub-modules: a state transition model, an observation model, an observation noise model, and a sensor model. The entire framework is trained in an end-to-end manner via a mean squared error (MSE) loss between the ground truth state $\hat{{\bf x}}_{t|t}$ and the estimated state ${\bf \bar{x}}_{t|t}$ at every timestep. We also supervise the intermediate modules via loss gradients $\mathcal{L}_{f_{\pmb {\theta}}}$ and $\mathcal{L}_{s_{\pmb {\xi}}}$. Given ground truth at time $t$, we apply the MSE loss gradient calculated between $\hat{{\bf x}}_{t|t}$ and the output of the state transition model to $f_{\pmb {\theta}}$ as in Eq.~\ref{eq:loss1}. We apply the intermediate loss gradients computed based on the ground truth observation $\hat{{\bf y}_t}$ and the output of the stochastic sensor model $\tilde{{\bf y}}_t$: 
    \begin{align}
    \label{eq:loss1}
    \mathcal{L}_{f_{\pmb {\theta}}} =  \| {\bf \bar{x}}_{t|t-1} - \hat{{\bf x}}_{t|t}\|_2^2,\ \ 
        \mathcal{L}_{s_{\pmb {\xi}}} =\| \tilde{{\bf y}_t} -  \hat{{\bf y}_t}\|_2^2.
    \end{align}
All models in the experiments were trained for 50 epochs with batch size 64, and a learning rate of $\eta = 10^{-5}$. We chose the model with the best performance on a validation set for testing. The ensemble size of the DEnKF was set to \textbf{32 ensemble members.}

\subsection{Visual Odometry Task}\label{KITTI}
In this experiment, we investigate performance on the popular KITTI Visual Odometry dataset~\cite{geiger2012we}. Following the same evaluation procedure as our baselines~\cite{kloss2021train, jonschkowski2018differentiable, haarnoja2016backprop}, 
we define the state of the moving vehicle as a 5-dimensional vector ${\bf x} = [x, y, \theta, v, \dot{\theta}]^T$, including the position and orientation of the vehicle, and the linear and angular velocity w.r.t.~the current heading direction $\theta$. The raw observation $\bf y$ corresponds to the RGB camera image of the current frame and a difference image between the current frame and the previous frame, where ${\bf y} \in \mathbb{R}^{150 \times 50 \times 6}$. The learned observation $\tilde{{\bf y}}$ is defined as $\tilde{{\bf y}} = [v, \dot{\theta}]^T$, since only the relative changes of position and orientation can be captured between two frames.

\textbf{Data:} The KITTI Visual Odometry dataset consists of 11 trajectories with ground truth pose (translation and rotation matrices) of a vehicle driving in urban areas with a data collection rate around 10Hz. To facilitate learning, we standardize the data on every dimension to have a 0 mean and a standard deviation of 1 during training.

\begin{table*}[h]
\caption{Result evaluations on KITTI Visual Odometry task measured in RMSE and MAE. m/m and def/m denote the translational error and the rotational error. Results for dEKF, dPF, and dPF-M-lrn are reproduced for detailed comparisons.}
\label{Tab:result_KITTI}
\begin{center}
\scalebox{1}{
\begin{tabular}{c c c c c c c c}
    \toprule
     \multirow{2}{3em}{Method} &\multirow{2}{3em}{RMSE}
     &\multirow{2}{3em}{MAE} 
     &\multicolumn{2}{c}{Test 100}
     &\multicolumn{2}{c}{Test 100/200/400/800}
     & \multirow{2}{5em}{Wall clock time (s)}\\
     & &
     &m/m 
     &deg/m
     &m/m
     &deg/m
     &
     \\
    \midrule
     dEKF~\cite{kloss2021train}
     &2.9366$\pm$0.006 &1.9738$\pm$0.027
     &0.2646$\pm$0.004 &0.1386$\pm$0.002
     &0.3159$\pm$0.002 &0.0923$\pm$0.005
     &\bf 0.0463$\pm$0.004
     \\
    DPF~\cite{jonschkowski2018differentiable}
     &2.2575$\pm$0.027
     &1.4671$\pm$0.014
     &0.1344$\pm$0.002  
     &0.1203$\pm$0.007
     &0.2255$\pm$0.001 
     &0.0716$\pm$0.004
     &0.0486$\pm$0.005
     \\
    dPF-M-lrn~\cite{kloss2021train}
     &2.1001$\pm$0.012
     &1.4016$\pm$0.004
     &0.1720$\pm$0.010 
     &0.0974$\pm$0.009
     &0.1848$\pm$0.004
     &0.0611$\pm$0.003
     &0.0693$\pm$0.011
     \\
    DEnKF
     &\bf 1.3329$\pm$0.043
     &\bf 1.1349$\pm$0.024
     &\bf 0.0249$\pm$0.001 
     &\bf 0.0506$\pm$0.001
     &\bf 0.0460$\pm$0.002
     &\bf 0.0353$\pm$0.001
     &0.0603$\pm$0.001
     \\
\bottomrule
\multicolumn{8}{l}{Means$\pm$standard errors.} \\
\end{tabular}}
\end{center}
\vspace{-0.2in}
\end{table*}

\begin{figure}[h!]
\centering
\includegraphics[width=\linewidth]{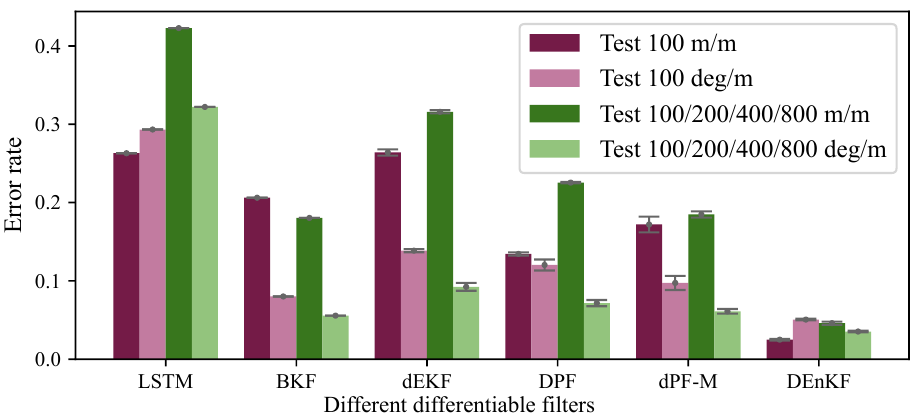}
\caption{Visual Odometry results with different differentiable filters: the error rate for LSTM and BKF are reported from~\cite{haarnoja2016backprop}, dEKF, DPF, and dPF-M are reproduced.}
\label{fig:KITTI}
\vspace{-0.1in}
\end{figure} 

\textbf{Results:} We assess the performance of state estimation using an 11-fold cross-validation withholding 1 trajectory at each time. We report the root mean squared error (RMSE), mean absolute error (MAE), and the standard KITTI benchmark metrics, the translational error (m/m), and the rotational error (deg/m) in Table~\ref{Tab:result_KITTI}. The error metrics are computed from the test trajectory over all subsequences of 100 timesteps, and all subsequences of 100, 200, 400, and 800 timesteps. Figure~\ref{fig:KITTI} shows the performance of DEnKF and other differentiable filtering techniques. Note that lower error metrics can be obtained by imposing domain- and data-specific information, i.e., using stereo images~\cite{lenac2018exactly}, incorporating LiDAR~\cite{zhang2015visual,chou2021efficient}, or applying SLAM and loop-closure related assumptions~\cite{lenac2018exactly,cvivsic2018soft}. However, we opt for the most commonly used setup when comparing filtering technique in a task-agnostic fashion (as performed in~\cite{kloss2021train, jonschkowski2018differentiable, haarnoja2016backprop}) to ensure fair and comparable evaluations. 

\textbf{Comparison:} Table~\ref{Tab:result_KITTI} presents the outcomes of our proposed method in comparison with the existing state-of-the-art differentiable filters, including differentiable Extended Kalman filter (dEKF)~\cite{kloss2021train}, differentiable particle filter (DPF)~\cite{jonschkowski2018differentiable}, and modified differentiable particle filter with learned process and process noise models (dPF-M-lrn)~\cite{kloss2021train}. To provide a fair comparison, we do not include unstructured LSTM models as baselines since prior works~\cite{haarnoja2016backprop,kloss2021train} have shown that LSTM models do not achieve comparable results. In our comparison, we use the same pre-trained sensor model $s_{\pmb {\xi}}$ with the same visual inputs and integrate it into all the DF frameworks evaluated here. In this experiment, the motion model of the vehicle is known. The only unknown part of the state is the velocities. Therefore, we use the learnable process model to update those state variables and use the known motion model to update the ($x,y, \theta$). For dEKF, we supply the computed Jacobian matrix in training and testing since the motion model is known. For DPF, we use 100 particles to train and test. DPF contains a different learnable module called observation likelihood estimation model $l$, which takes an image embedding and outputs a likelihood for updating each particle's weight. 

Table~\ref{Tab:result_KITTI} shows that DEnKF achieves a RMSE of 1.33, which is $\thicksim$54\%, $\thicksim$41\%, and $\thicksim$37\% lower than that of dEKF, DPF, and dPF-M-lrn, respectively. 
Specifically, DEnKF reduces the translational error by $\thicksim$85\%, $\thicksim$79\%, and $\thicksim$75\% for Test 100/200/400/800 compared to dEKF, DPF, and dPF-M-lrn. It also reduces the rotational error by $\thicksim$62\%, $\thicksim$51\%, and $\thicksim$42\% for each baseline.
Notably, dPF-M-lrn manifests the best performance among all the baselines, it implements learnable process noise model as described in Eq.~\ref{eq:04}, and it uses a Gaussian Mixture Model for computing the likelihood for all particles. 
While dPF-M-lrn performs the best among all baselines, DEnKF shows higher tracking accuracy than dPF-M-lrn and runs 0.009s faster. It is worth noting that the inference time for dPF-M-lrn is higher than any other DF in Table~\ref{Tab:result_KITTI}.
\begin{figure}[h!]
\centering
\includegraphics[width=\linewidth]{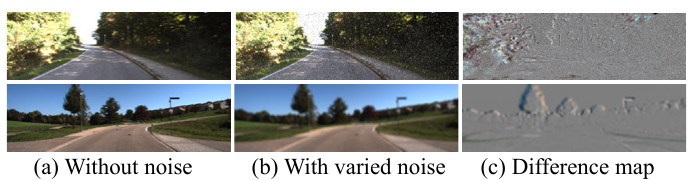}
\caption{Different visual inputs during testing. (a) the raw observation; (b) the visual inputs with 2 types of noise: salt-and-pepper and blurring; (c) the difference map.}
\label{fig:KITTI_noise}
\vspace{-0.1in}
\end{figure} 

\textbf{Noisy and missing observation:} According to~\cite{ceccarelli2022rgb}, failures
of vehicle cameras may compromise autonomous driving performance -- potentially even leading to injuries and death. Common failures are listed by~\cite{ceccarelli2022rgb}, i.e., brightness, blurred vision, and brackish. 
We conducted an evaluation of the performance of DEnKF and other DFs in the presence of noisy observations during inference. In Fig.\ref{fig:KITTI_noise}, we added salt-and-pepper and blurring effects to the test images, and reported the performance of DFs under these conditions in Table\ref{Tab:noise_KITTI} (top and mid). Our findings show that DEnKF with noise performs worse compared to DEnKF without noise, with a 17\% and 29\% increase in translational and rotational error, respectively, compared to the metrics from Table~\ref{Tab:result_KITTI}. However, DEnKF remains more robust against noise perturbations than dEKF, DPF, and dPF-M-lrn, achieving $\thicksim$80\%, $\thicksim$66\%, and $\thicksim$59\% improvement on translational error with salt-and-pepper noise for Test 100/200/400/800. We also performed an experiment on missing observations by providing no visual input with a chance of 30\% at every timestep. In this case, DEnKF's modularity allows the state transition model $f_{\pmb{\theta}}$ to propagate the state forward through the state space, whereas other RNN-based filters remain in the same hidden state until an observation is processed. Error metrics for this scenario are reported in Table~\ref{Tab:noise_KITTI} (bottom), where we re-built the process model for each DFs to account for such problems. DEnKF's incorporation of a stochastic neural network in the forward model handling missing observation scenarios outperforms other DFs, resulting in a reduction of $\thicksim$54\% and $\thicksim$36\% for translational and rotational error, respectively, compared to dPF-M-lrn.

\begin{table}[t]
\caption{Result evaluations on different types of noisy visual input and with 30\% missing observation.}
\label{Tab:noise_KITTI}
\begin{center}
\scalebox{1}{
\begin{tabular}{c c c c c c}
    \toprule
    &
     &\multicolumn{2}{c}{Test 100}
     &\multicolumn{2}{c}{Test 100/200/400/800}
     \\
     &
     &m/m
     &deg/m
     &m/m
     &deg/m
     \\
    \midrule
    \multirow{4}{*}{\STAB{\rotatebox[origin=c]{90}{S\&P Noise}}}
     &dEKF
     &0.267 &0.094
     &0.364 &0.061
     \\
    & DPF
     &0.249 &0.159
     &0.296 &0.052
     \\
    & dPF-M-lrn
     &0.207 &0.090
     &0.246 &0.048
     \\
    & DEnKF
     &\bf 0.029
     &\bf 0.089
     &\bf 0.102 
     &\bf 0.039
     \\
     \midrule
         \multirow{4}{*}{\STAB{\rotatebox[origin=c]{90}{Blurring}}}
     &dEKF
     &0.235 &0.137
     &0.390 &0.104
     \\
    & DPF
     &0.144 &0.122
     &0.236 &0.101
     \\
    & dPF-M-lrn
     &0.141 &0.073
     &0.172 &0.085
     \\
    & DEnKF
     &\bf 0.028
     &\bf 0.064
     &\bf 0.054 
     &\bf 0.050
     \\
    \midrule
    \multirow{4}{*}{\STAB{\rotatebox[origin=c]{90}{Missing}}}
     &dEKF
     &0.254 &0.173
     &0.298 &0.134
     \\
    & DPF
     &0.245 &0.165
     &0.273 &0.131
     \\
    & dPF-M-lrn
     &0.204 &0.116
     &0.230 &0.093
     \\
    & DEnKF
     &\bf 0.112
     &\bf 0.089
     &\bf 0.188
     &\bf 0.057
     \\
\bottomrule
\end{tabular}}
\end{center}
\vspace{-0.2in}
\end{table}

\begin{figure}[b!]
\centering
\includegraphics[width=0.9\linewidth]{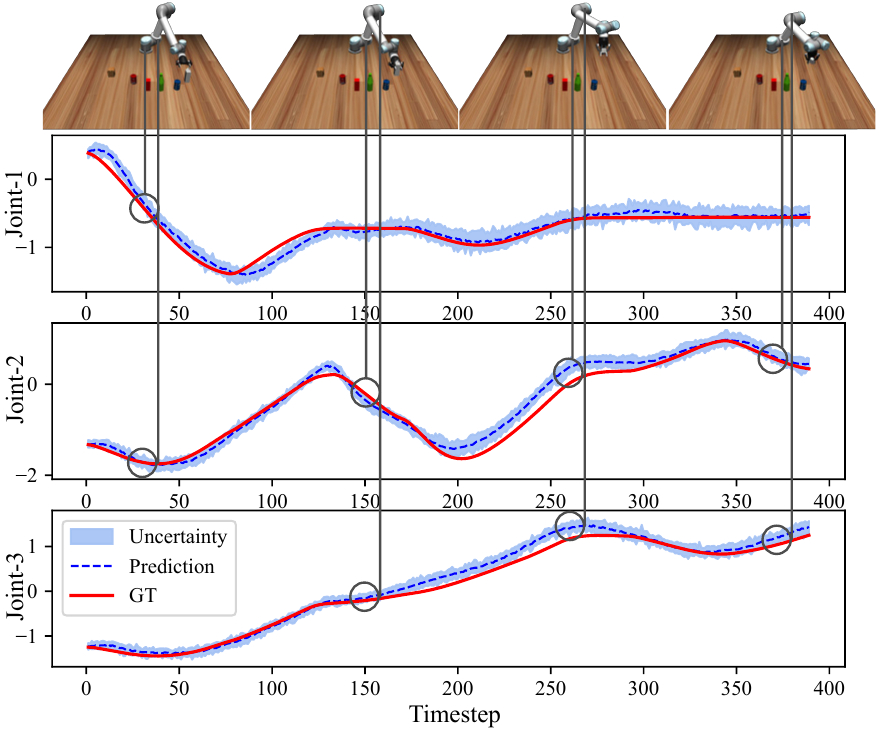}
\caption{Results on UR5 manipulation in simulation. Top: the robot joint configuration at varied timestep; Bottom: 3 key joint angle trajectory; The vertical lines mapping from the top to the bottom align the tracking results of the exact joint angle in time for each key joint.}
\label{fig:UR5_sim}
\end{figure} 
\begin{table*}[t!]
\caption{Result evaluations on UR5 manipulation task measured in MAE from 3 different domains -- real-world, simulation, and sim-to-real. Results for dEKF, dPF, and dPF-M-lrn are reproduced for detailed comparisons.}
\label{Tab:result_manipulation}
\begin{center}
\scalebox{1}{
\begin{tabular}{c c c c c c c c}
    \toprule
     \multirow{2}{3em}{Method} 
     &\multicolumn{2}{c}{Real-world}
     &\multicolumn{2}{c}{Simulation}
     &\multicolumn{2}{c}{Sim-to-real}
     &\multirow{2}{5em}{Wall clock time (s)}
     \\
     &Joint (deg)  &EE (cm)
     &Joint (deg)  &EE (cm)
     &Joint (deg)  &EE (cm)
     \\
    \midrule
     dEKF~\cite{kloss2021train}
     & 16.0862$\pm$0.063
     & 5.6680$\pm$0.060
     & 4.9357$\pm$0.224
     & 1.9112$\pm$0.148
     & 8.3041$\pm$0.525
     & 4.3645$\pm$0.072
     & \bf 0.0469$\pm$0.003
     \\
    DPF~\cite{jonschkowski2018differentiable}
     & 15.9302$\pm$0.080
     & 5.0834$\pm$0.301
     & 4.4623$\pm$0.220
     & 1.5135$\pm$0.191
     & 5.9531$\pm$0.031
     & 3.9695$\pm$0.006
     & 0.0515$\pm$0.002
     \\
    dPF-M-lrn~\cite{kloss2021train}
     & 12.8366$\pm$0.086
     & 3.9521$\pm$0.436
     & 3.8233$\pm$0.230
     & 1.2639$\pm$0.081
     & 5.4389$\pm$0.011
     & 3.9405$\pm$0.014
     & 0.0854$\pm$0.001
     \\
    DEnKF
     &\bf 11.4222$\pm$0.005
     &\bf 3.4260$\pm$0.002
     &\bf 2.5587$\pm$0.093
     &\bf 0.8241$\pm$0.019
     &\bf 3.9531$\pm$0.034
     &\bf 2.6368$\pm$0.002
     & 0.0712$\pm$0.002
     \\
\bottomrule
\multicolumn{8}{l}{Means$\pm$standard errors.} \\
\end{tabular}}
\end{center}
\vspace{-0.2in}
\end{table*}
\subsection{Robot Manipulation Task}
In the second experiment, we assess the efficacy of DEnKF in a challenging robot manipulation setting. Specifically, we train and employ DEnKF to monitor the state of a UR5 robot while performing tabletop arrangement tasks. Similar to behavioral cloning from observation tasks~\cite{torabi2018behavioral}, actions are not provided in this experiment, and the DEnKF is trained to learn to propagate state over time. The robot state is defined as ${\bf x} = [J_1, \cdots, J_7, x, y, z]^T$, where $J_1$-$J_7$ denote the 7 joint angle of the UR5 robot, and $(x,y,z)$ represents the 3D robot end-effector (EE) position w.r.t.~$(0,0,0)$ which is the center of the manipulation platform. 
As shown in Fig.~\ref{fig:UR5_sim}(top), raw observations ${\bf y} \in \mathbb{R}^{224 \times 224 \times 3}$ are images captured from a camera placed in front of the table. The learned observation $\tilde{{\bf y}}$ is defined to have the same dimension as the robot state, where $\tilde{{\bf y}} = [J_1, \cdots, J_7, x, y, z]^T$.

\textbf{Data:} The data collection process is conducted both in the MuJoCo~\cite{todorov2012mujoco} simulator and in the real-world. We record the UR5 robot operating on a random object by performing one of ``pick'', ``push'', and ``put down'' actions. We collect 2,000 demonstrations in simulation and 100 on the real robot, changing the location of each object for each demonstration. We use ABR control and robosuite~\cite{robosuite2020} in addition to MuJoCo to ensure rigorous dynamics in the simulator. Each sequence length is around 350 steps with 0.08 sec as the timestep. We use an 80/20 data split for training and testing.

\begin{figure}[t]
\centering
\includegraphics[width=\linewidth]{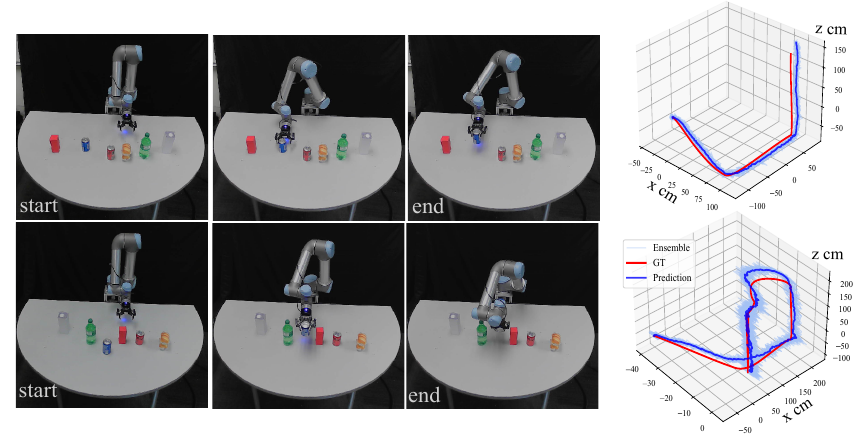}
\caption{EE positions visualization. Top: UR5 executes ``pick up'' action; Bottom: UR5 executes ``put down'' action.}
\label{fig:UR5_EE}
\vspace{-0.1in}
\end{figure}

\textbf{Result:} We conduct a performance evaluation of DEnKF in three different domains, namely real-world, simulation, and sim-to-real. We train two separate DEnKFs on simulation and real-world datasets, respectively, and then perform sim-to-real transfer by fine-tuning the simulation-trained DEnKF on real-world data. State estimation with uncertainty measurement using distributed ensemble members in simulation is illustrated in Fig.~\ref{fig:UR5_sim}. Following the same comparison protocol as in Sec.~\ref{KITTI}, we supply all DFs the same pre-trained sensor model $s_{\pmb {\xi}}$ in each domain, but no known motion model is enabled at this time. We train all learnable modules except $s_{\pmb {\xi}}$ and reported the mean absolute error (MAE) in the joint angle space (deg) and end-effector positions (cm) for DEnKFs and other DF baselines.
The experimental results indicate that DEnKF and the other baseline DFs are capable of achieving domain adaptation by fine-tuning the simulation framework for real-world scenarios.  Notably, the DEnKF with sim-to-real transfer achieves accurate state estimation, resulting in a reduction of 29\% in MAE for joint angle space and 33\% for end-effector (EE) positions when compared to the dPF-M-lrn. In Table~\ref{Tab:result_manipulation}, the DEnKF with sim-to-real transfer exhibits an average of 2.6cm offset (MAE) from the ground truth for EE positions across testing sequences. We further analyze the state tracking in EE space by visualizing the EE trajectories in 3D, as depicted in Fig.~\ref{fig:UR5_EE}, where the fine-tuned DEnKF is utilized to estimate the state with two real-robot test examples of action sequence ``pick up" and ``put down".

\begin{figure}[t]
\centering
\includegraphics[width=\linewidth]{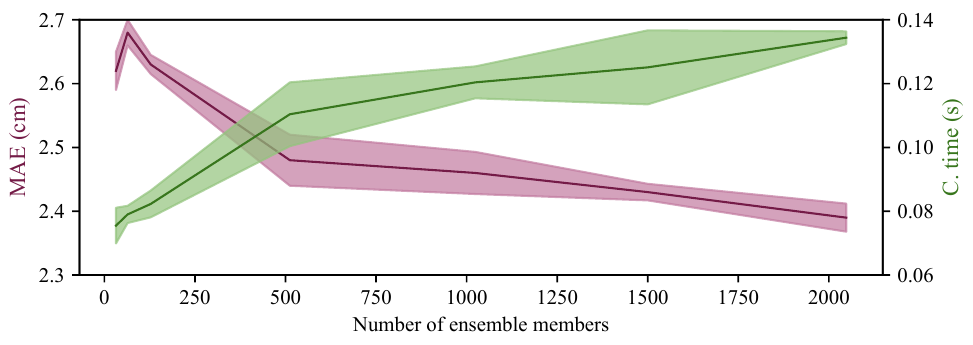}
\caption{Computational time vs. MAE (EE position) with increasing number of ensemble members.}
\label{fig:UR5_time}
\vspace{-0.2in}
\end{figure}

Moreover, an additional experiment is conducted to test the trade-off between the accuracy and computational performance of proposed DEnKF framework as shown in Fig.~\ref{fig:UR5_time}. Tellingly, increasing the number of ensemble members can have a substantial positive effect on performance ($+9.6\%$) while the  computational overhead is only marginally affected (increase from 0.075 sec to 0.134 sec). 

\section{Conclusions}
In this paper, we present the Differentiable Ensemble Kalman Filters (DEnKF) as an extended version of Ensemble Kalman Filters, and demonstrate their applications to state estimation tasks. We show that our framework is applicable in both simulation and real-world, and that it is capable of performing state estimation with complex tasks, e.g., KITTI visual odometry, and robot manipulation. We also discuss state tracking in high-dimensional observation spaces with varied noise conditions and missing observations. In particular, DEnKF manages to decrease the error metrics on translation and rotation by at least 59\% and 36\% with noisy observations versus the state-of-the-art approaches.
These experiments manifest the DEnKF significantly improves the tracking accuracy and uncertainty estimates, thus, has great potential in many robotic applications.

The proposed framework is modular, which allows for flexibility in using individual components separately. However, it should be noted that various learning tasks may require distinct curricula. For example, challenging visual tasks may necessitate an extended training period for the sensor model before it can be incorporated into end-to-end learning. Therefore, a universal curriculum that guarantees optimal performance of all sub-modules in every situation does not currently exist. In future research, we plan to explore the potential of the DEnKF framework in detecting perturbations as a downstream application. Specifically, leveraging the learned system dynamics from the state transition model and the mapping from observation to state space learned by the sensor model and the Kalman update step, we aim to use the distance between the outputs of the two steps to detect perturbations in the system. Overall, the results of this study suggest that DEnKF has great potential in many robotic applications.

\section*{Acknowledgment}
The authors gratefully acknowledge support of this work through a grant by “The Global KAITEKI Center” (TGKC) of the Global Futures Laboratory at Arizona State University. TGKC is a research alliance between Arizona State University and The KAITEKI Institute, an affiliate of the Mitsubishi Chemical Group.

\bibliographystyle{IEEEtran}
\bibliography{references}

\begin{thebibliography}{10}
\providecommand{\url}[1]{#1}
\csname url@rmstyle\endcsname
\providecommand{\newblock}{\relax}
\providecommand{\bibinfo}[2]{#2}
\providecommand\BIBentrySTDinterwordspacing{\spaceskip=0pt\relax}
\providecommand\BIBentryALTinterwordstretchfactor{4}
\providecommand\BIBentryALTinterwordspacing{\spaceskip=\fontdimen2\font plus
\BIBentryALTinterwordstretchfactor\fontdimen3\font minus
  \fontdimen4\font\relax}
\providecommand\BIBforeignlanguage[2]{{%
\expandafter\ifx\csname l@#1\endcsname\relax
\typeout{** WARNING: IEEEtran.bst: No hyphenation pattern has been}%
\typeout{** loaded for the language `#1'. Using the pattern for}%
\typeout{** the default language instead.}%
\else
\language=\csname l@#1\endcsname
\fi
#2}}

\bibitem{thrun2005probabilistic}
S.~Thrun, W.~Burgard, and D.~Fox, \emph{Probabilistic robotics}.\hskip 1em plus
  0.5em minus 0.4em\relax Cambridge, Mass.: MIT Press, 2005.

\bibitem{WANG2022102310}
L.~Wang, G.~Wang, S.~Jia, A.~Turner, and S.~Ratchev, ``Imitation learning for
  coordinated human–robot collaboration based on hidden state-space models,''
  \emph{Robotics and Computer-Integrated Manufacturing}, vol.~76, p. 102310,
  2022.

\bibitem{chen2011kalman}
S.~Chen, ``Kalman filter for robot vision: a survey,'' \emph{IEEE Transactions
  on industrial electronics}, vol.~59, no.~11, pp. 4409--4420, 2011.

\bibitem{reher2019dynamic}
J.~Reher, W.-L. Ma, and A.~D. Ames, ``Dynamic walking with compliance on a
  cassie bipedal robot,'' in \emph{2019 18th European Control Conference
  (ECC)}.\hskip 1em plus 0.5em minus 0.4em\relax IEEE, 2019, pp. 2589--2595.

\bibitem{thrun2002probabilistic}
S.~Thrun, ``Probabilistic robotics,'' \emph{Communications of the ACM},
  vol.~45, no.~3, pp. 52--57, 2002.

\bibitem{NEURIPS2018_5cf68969}
S.~S. Rangapuram, M.~W. Seeger, J.~Gasthaus, L.~Stella, Y.~Wang, and
  T.~Januschowski, ``Deep state space models for time series forecasting,'' in
  \emph{Advances in Neural Information Processing Systems}, S.~Bengio,
  H.~Wallach, H.~Larochelle, K.~Grauman, N.~Cesa-Bianchi, and R.~Garnett, Eds.,
  vol.~31.\hskip 1em plus 0.5em minus 0.4em\relax Curran Associates, Inc.,
  2018.

\bibitem{klushyn2021latent}
A.~Klushyn, R.~Kurle, M.~Soelch, B.~Cseke, and P.~van~der Smagt, ``Latent
  matters: Learning deep state-space models,'' \emph{Advances in Neural
  Information Processing Systems}, vol.~34, 2021.

\bibitem{kloss2021train}
A.~Kloss, G.~Martius, and J.~Bohg, ``How to train your differentiable filter,''
  \emph{Autonomous Robots}, pp. 1--18, 2021.

\bibitem{jonschkowski2018differentiable}
R.~Jonschkowski, D.~Rastogi, and O.~Brock, ``Differentiable particle filters:
  End-to-end learning with algorithmic priors,'' \emph{arXiv preprint
  arXiv:1805.11122}, 2018.

\bibitem{Sorenson}
H.~Sorenson, \emph{Kalman Filtering: Theory and Application}.\hskip 1em plus
  0.5em minus 0.4em\relax IEEE Press, Los Alamitos, 1985.

\bibitem{van2004sigma}
R.~Van Der~Merwe, \emph{Sigma-point Kalman filters for probabilistic inference
  in dynamic state-space models}.\hskip 1em plus 0.5em minus 0.4em\relax Oregon
  Health \& Science University, 2004.

\bibitem{haarnoja2016backprop}
T.~Haarnoja, A.~Ajay, S.~Levine, and P.~Abbeel, ``Backprop kf: Learning
  discriminative deterministic state estimators,'' in \emph{Advances in neural
  information processing systems}, 2016, pp. 4376--4384.

\bibitem{karkus2019differentiable}
P.~Karkus, X.~Ma, D.~Hsu, L.~P. Kaelbling, W.~S. Lee, and T.~Lozano-P{\'e}rez,
  ``Differentiable algorithm networks for composable robot learning,''
  \emph{arXiv preprint arXiv:1905.11602}, 2019.

\bibitem{corenflos2021differentiable}
A.~Corenflos, J.~Thornton, G.~Deligiannidis, and A.~Doucet, ``Differentiable
  particle filtering via entropy-regularized optimal transport,'' in
  \emph{International Conference on Machine Learning}.\hskip 1em plus 0.5em
  minus 0.4em\relax PMLR, 2021, pp. 2100--2111.

\bibitem{chen2021differentiable}
X.~Chen, H.~Wen, and Y.~Li, ``Differentiable particle filters through
  conditional normalizing flow,'' in \emph{2021 IEEE 24th International
  Conference on Information Fusion (FUSION)}.\hskip 1em plus 0.5em minus
  0.4em\relax IEEE, 2021, pp. 1--6.

\bibitem{wang2019dualsmc}
Y.~Wang, B.~Liu, J.~Wu, Y.~Zhu, S.~S. Du, L.~Fei-Fei, and J.~B. Tenenbaum,
  ``Dualsmc: Tunneling differentiable filtering and planning under continuous
  pomdps,'' \emph{arXiv preprint arXiv:1909.13003}, 2019.

\bibitem{wen2021end}
H.~Wen, X.~Chen, G.~Papagiannis, C.~Hu, and Y.~Li, ``End-to-end semi-supervised
  learning for differentiable particle filters,'' in \emph{2021 IEEE
  International Conference on Robotics and Automation (ICRA)}.\hskip 1em plus
  0.5em minus 0.4em\relax IEEE, 2021, pp. 5825--5831.

\bibitem{lee2020multimodal}
M.~A. Lee, B.~Yi, R.~Mart{\'\i}n-Mart{\'\i}n, S.~Savarese, and J.~Bohg,
  ``Multimodal sensor fusion with differentiable filters,'' in \emph{2020
  IEEE/RSJ International Conference on Intelligent Robots and Systems
  (IROS)}.\hskip 1em plus 0.5em minus 0.4em\relax IEEE, 2020, pp.
  10\,444--10\,451.

\bibitem{wagstaff2022self}
B.~Wagstaff, E.~Wise, and J.~Kelly, ``A self-supervised, differentiable kalman
  filter for uncertainty-aware visual-inertial odometry,'' in \emph{2022
  IEEE/ASME International Conference on Advanced Intelligent Mechatronics
  (AIM)}.\hskip 1em plus 0.5em minus 0.4em\relax IEEE, 2022, pp. 1388--1395.

\bibitem{evensen2003ensemble}
G.~Evensen, ``The ensemble kalman filter: Theoretical formulation and practical
  implementation,'' \emph{Ocean dynamics}, vol.~53, no.~4, pp. 343--367, 2003.

\bibitem{roth2017ensemble}
M.~Roth, G.~Hendeby, C.~Fritsche, and F.~Gustafsson, ``The ensemble kalman
  filter: a signal processing perspective,'' \emph{EURASIP Journal on Advances
  in Signal Processing}, vol. 2017, no.~1, pp. 1--16, 2017.

\bibitem{houtekamer2016review}
P.~L. Houtekamer and F.~Zhang, ``Review of the ensemble kalman filter for
  atmospheric data assimilation,'' \emph{Monthly Weather Review}, vol. 144,
  no.~12, pp. 4489--4532, 2016.

\bibitem{jospin2022}
L.~V. Jospin, H.~Laga, F.~Boussaid, W.~Buntine, and M.~Bennamoun, ``Hands-on
  bayesian neural networks—a tutorial for deep learning users,'' \emph{IEEE
  Computational Intelligence Magazine}, vol.~17, no.~2, pp. 29--48, 2022.

\bibitem{lakshminarayanan2017simple}
B.~Lakshminarayanan, A.~Pritzel, and C.~Blundell, ``Simple and scalable
  predictive uncertainty estimation using deep ensembles,'' \emph{Advances in
  neural information processing systems}, vol.~30, 2017.

\bibitem{chua2018deep}
K.~Chua, R.~Calandra, R.~McAllister, and S.~Levine, ``Deep reinforcement
  learning in a handful of trials using probabilistic dynamics models,''
  \emph{Advances in neural information processing systems}, vol.~31, 2018.

\bibitem{chen2021auto}
Y.~Chen, D.~Sanz-Alonso, and R.~Willett, ``Auto-differentiable ensemble kalman
  filters,'' \emph{arXiv preprint arXiv:2107.07687}, 2021.

\bibitem{zhang2007system}
S.~Zhang, M.~Harrison, A.~Rosati, and A.~Wittenberg, ``System design and
  evaluation of coupled ensemble data assimilation for global oceanic climate
  studies,'' \emph{Monthly Weather Review}, vol. 135, no.~10, pp. 3541--3564,
  2007.

\bibitem{quang2016high}
P.~B. Quang and V.~Tran, ``High-dimensional simulation experiments with
  particle filter and ensemble kalman filter,'' \emph{Applied Mathematics in
  Engineering and Reliability}, p.~1, 2016.

\bibitem{gal2016dropout}
Y.~Gal and Z.~Ghahramani, ``Dropout as a bayesian approximation: Representing
  model uncertainty in deep learning,'' in \emph{international conference on
  machine learning}.\hskip 1em plus 0.5em minus 0.4em\relax PMLR, 2016, pp.
  1050--1059.

\bibitem{geiger2012we}
A.~Geiger, P.~Lenz, and R.~Urtasun, ``Are we ready for autonomous driving? the
  kitti vision benchmark suite,'' in \emph{2012 IEEE conference on computer
  vision and pattern recognition}.\hskip 1em plus 0.5em minus 0.4em\relax IEEE,
  2012, pp. 3354--3361.

\bibitem{lenac2018exactly}
K.~Lenac, J.~{\'C}esi{\'c}, I.~Markovi{\'c}, and I.~Petrovi{\'c}, ``Exactly
  sparse delayed state filter on lie groups for long-term pose graph slam,''
  \emph{The International Journal of Robotics Research}, vol.~37, no.~6, pp.
  585--610, 2018.

\bibitem{zhang2015visual}
J.~Zhang and S.~Singh, ``Visual-lidar odometry and mapping: Low-drift, robust,
  and fast,'' in \emph{2015 IEEE International Conference on Robotics and
  Automation (ICRA)}.\hskip 1em plus 0.5em minus 0.4em\relax IEEE, 2015, pp.
  2174--2181.

\bibitem{chou2021efficient}
C.-C. Chou and C.-F. Chou, ``Efficient and accurate tightly-coupled
  visual-lidar slam,'' \emph{IEEE Transactions on Intelligent Transportation
  Systems}, 2021.

\bibitem{cvivsic2018soft}
I.~Cvi{\v{s}}i{\'c}, J.~{\'C}esi{\'c}, I.~Markovi{\'c}, and I.~Petrovi{\'c},
  ``Soft-slam: Computationally efficient stereo visual simultaneous
  localization and mapping for autonomous unmanned aerial vehicles,''
  \emph{Journal of field robotics}, vol.~35, no.~4, pp. 578--595, 2018.

\bibitem{ceccarelli2022rgb}
A.~Ceccarelli and F.~Secci, ``Rgb cameras failures and their effects in
  autonomous driving applications,'' \emph{IEEE Transactions on Dependable and
  Secure Computing}, 2022.

\bibitem{torabi2018behavioral}
F.~Torabi, G.~Warnell, and P.~Stone, ``Behavioral cloning from observation,''
  \emph{arXiv preprint arXiv:1805.01954}, 2018.

\bibitem{todorov2012mujoco}
E.~Todorov, T.~Erez, and Y.~Tassa, ``Mujoco: A physics engine for model-based
  control,'' in \emph{2012 IEEE/RSJ International Conference on Intelligent
  Robots and Systems}.\hskip 1em plus 0.5em minus 0.4em\relax IEEE, 2012, pp.
  5026--5033.

\bibitem{robosuite2020}
Y.~Zhu, J.~Wong, A.~Mandlekar, and R.~Mart\'{i}n-Mart\'{i}n, ``robosuite: A
  modular simulation framework and benchmark for robot learning,'' in
  \emph{arXiv preprint arXiv:2009.12293}, 2020.

\bibitem{butala2008asymptotic}
M.~D. Butala, J.~Yun, Y.~Chen, R.~A. Frazin, and F.~Kamalabadi, ``Asymptotic
  convergence of the ensemble kalman filter,'' in \emph{2008 15th IEEE
  International Conference on Image Processing}.\hskip 1em plus 0.5em minus
  0.4em\relax IEEE, 2008, pp. 825--828.

\bibitem{kasanicky2017ensemble}
I.~Kasanick{\`y}, ``Ensemble kalman filter on high and infinite dimensional
  spaces,'' \emph{Univerzita Karlova, Matematicko-fyzik{\'a}ln{\'\i} fakulta},
  2017.

\bibitem{mandel2006efficient}
J.~Mandel, \emph{Efficient implementation of the ensemble Kalman filter}.\hskip
  1em plus 0.5em minus 0.4em\relax University of Colorado at Denver and Health
  Sciences Center, 2006.

\bibitem{hommels2009comparison}
A.~Hommels, A.~Murakami, and S.-I. Nishimura, ``A comparison of the ensemble
  kalman filter with the unscented kalman filter: application to the
  construction of a road embankment,'' \emph{Geotechniek}, vol.~13, no.~1,
  p.~52, 2009.

\bibitem{mandel2011convergence}
J.~Mandel, L.~Cobb, and J.~D. Beezley, ``On the convergence of the ensemble
  kalman filter,'' \emph{Applications of Mathematics}, vol.~56, no.~6, pp.
  533--541, 2011.

\bibitem{cybenko1989approximation}
G.~Cybenko, ``Approximation by superpositions of a sigmoidal function,''
  \emph{Mathematics of control, signals and systems}, vol.~2, no.~4, pp.
  303--314, 1989.

\bibitem{pu2009ensemble}
Z.~Pu and J.~Hacker, ``Ensemble-based kalman filters in strongly nonlinear
  dynamics,'' \emph{Advances in atmospheric sciences}, vol.~26, pp. 373--380,
  2009.

\bibitem{bergou2019convergence}
E.~H. Bergou, S.~Gratton, and J.~Mandel, ``On the convergence of a non-linear
  ensemble kalman smoother,'' \emph{Applied Numerical Mathematics}, vol. 137,
  pp. 151--168, 2019.

\end{thebibliography}

\appendix[Theoretical properties of EnKF]
This section provides an overview of the theoretical underpinnings of Ensemble Kalman Filters (EnKF), outlines key differences between EnKF and Kalman Filters (KF), and highlights the significance of EnKF in practical applications.

\subsection{Asymptotic Convergence to KF}
Research in~\cite{butala2008asymptotic} has demonstrated that EnKF asymptotically converges to the optimal estimates provided by KF in a finite space. It is important to note that EnKF is a Monte Carlo algorithm that offers computationally tractable solutions to high-dimensional state estimation problems, which converge to KF~\cite{butala2008asymptotic}.

\subsection{Convergence in Hilbert Space}
The convergence of EnKF has been proven in~\cite{kasanicky2017ensemble} for infinite-dimensional state space models. Specifically, the proof shows that when a sequence of states ${\bf X}_{t}$ represents a system with stochastic dynamics defined on a Hilbert space and EnKF is applied, the empirical ensemble converges to the reference ensemble in $L^p$ for all $p \in [1,\infty)$. The convergence in the infinite dimensional Hilbert space is an important insight in machine learning research, as it offers a theoretical foundation for the use of EnKF in high-dimensional state estimation problems.


\subsection{Computational Performance}
EnKF has a computational complexity of around $O(E^2 n)$, where $E$ is the number of ensemble members and $n$ is the dimension of the state space, according to Mandel et al.~\cite{mandel2006efficient}. In contrast, the computational complexity of the Extended Kalman Filter (EKF), an alternative filtering approach for non-linear systems, is $O(n^3)$~\cite{thrun2005probabilistic}. One of the advantages of EnKF over EKF is that it does not require explicit construction of the covariance matrix since this information is implicitly captured by the ensemble. This reduces the computational and memory burden for large-scale estimation problems.

\subsection{Comparison with UKF}
The Unscented Kalman Filter (UKF) employs a set of sigma points that are propagated through the actual non-linear function. These points are selected to match the mean, covariance, and higher-order moments of a Gaussian random variable. In some cases, UKF has the tendency to overestimate the state compared to EnKF, as illustrated in a an example in~\cite{hommels2009comparison}. Still, both EnKF and UKF are effective state estimation methods for non-linear systems, and the choice of method should depend on the specific task at hand, in other words, the suitability of either method is determined by factors such as the complexity of the model, the availability of prior information, and the desired computational efficiency.

\subsection{Convergence of DEnKF}

Ensemble Kalman Filter (EnKF) are proven to converge in linear systems~\cite{mandel2011convergence}. In the context of our \emph{differentiable} Ensemble Kalman Filters (DEnKF), the prediction step is substituted with a stochastic neural network for the state transition function, as introduced in Sec.~\ref{sec:bnn}. $\bf H$ in Eq.~\ref{eq:2} is learned through multi-layer perceptrons as part of the observation model. In theory, a neural network can act as a function approximator capable of modeling exact linear functions, which has been demonstrated by the universal approximation theorem~\cite{cybenko1989approximation}. Hence, the proposed state transition function can be approximated to any degree of accuracy when the dynamics model is linear, and this holds true for the observation model in DEnKF as it approximates a linear process. Accordingly, the DEnKF satisfies the criteria necessary for the linear EnKF convergence proof. With regards to nonlinear systems, there is no global convergence proof for the EnKF. However, in practice it is often the preferred filtering approach in this setting since it empirically demonstrates strong performance in a variety of applications with strong non-linearities~\cite{pu2009ensemble}. In addition, recent work has shown that specific variants of the filter~\cite{bergou2019convergence}, i.e., the ensemble Kalman smoother, are proven to converge even in the nonlinear settings under certain assumptions. 

\end{document}